\documentclass{svproc}
\usepackage{url}

\usepackage{amsmath}
\usepackage{amsfonts}
\usepackage{amssymb}

\usepackage{cleveref}
\Crefname{equation}{Eq.}{Eqs.}
\Crefname{figure}{Fig.}{Figs.}
\Crefname{tabular}{Tab.}{Tabs.}
\Crefname{section}{Sect.}{Sects.}

\usepackage{graphicx}
\usepackage[caption=false]{subfig}

\def\R{{\mathbb{R}}}

\def\Rotx#1{\hbox{Rot}_x({#1})}

\def\Rotz#1{\hbox{Rot}_z({#1})}
\def\Transl{\hbox{Transl}}
\def\axis{\theta}

\begin{document}
\mainmatter              % start of a contribution
%
%\title{ Cuspidal 6R Robot with up to $16$ Solutions}
\title{Analysis of a Cuspidal 6R Robot}
\titlerunning{Analysis of a Cuspidal 6R Robot}
%\titlerunning{A Cuspidal 6R Robot with up to $16$ Solutions}
\author{Alexander Feeß \and Martin Weiß}
\authorrunning{Alexander Feeß, Martin Weiß} % abbreviated author list (for running head)
\tocauthor{Alexander Feeß and Martin Weiß}

\institute{OTH Regensburg, 93053 Regensburg, Germany,\\
\email{alexander.feess@oth-regensburg.de \\ martin.weiss@oth-regensburg.de}}

% WWW home page: \texttt{http://users/\homedir iekeland/web/welcome.html}

\maketitle   

\begin{abstract}
We present a theoretical and numerical analysis of the kinematics for the \emph{Transpressor}, a cuspidal $6$R robot. It admits up to $16$ inverse kinematics solutions which are described geometrically. For special target poses, we provide the solutions analytically, and present a simple numerical solver for the general case. Moreover, an estimation of the Jacobian determinant on a path between two solutions proves cuspidality for a class of robots similar to the transpressor. 

To the best of the authors' knowledge this is the first 6R robot where singularity free paths are provided for a wide class of solutions, without the need for numerical investigations. 

% We would like to encourage you to list your keywords within
% the abstract section using the \keywords{...} command.
\end{abstract}

\section{Introduction}

A general $6$R robot may have up to $16$ inverse kinematic solutions (IKS), ignoring joint limits. This was proven in \cite{E.J.FPrimrose.1986} and a first numerical example showing that the bound is sharp is given in \cite{RachidManseur.1989}. Most commercial robots have their last $3$ axes intersecting (spherical wrist) and a simple regional structure, resulting in $8$ solutions. 
% . Here the regional part of the kinematics has at most 4 solutions, the wrist has 2 solutions (see e.g. \cite[Chapter 5.1]{Selig.2005})
With the recent rise in popularity of cobots, new manipulators with differing structures appeared, including models with non-spherical wrists. These can have advantages in their construction, but are mathematically more challenging, as discussed in \cite{Salunkhe.2023,Elias.30.01.2025}. For some of these, the inverse kinematics problem can be solved analytically and might even still have only $8$ solutions like the Universal Robot URx series \cite{RyanKeating.2014,Yang.2023}. But generally, there are up to $16$ solutions, and no analytical solution is known e.g. for the FANUC CRX-10ia/L, Yaskawa HC10DTP and Kinova Link 6. 

One property that is expected of these models is \emph{cuspidality}: A non-singular motion from one IKS to another is possible. This effect has mostly been studied for $3$R robots, where a characterization of this property has been established and classification results exist, see \cite{Wenger.2019}. For $6$R manipulators, these are still open problems. Numerical investigations of specific robots include \cite{Carbonari.2023,Salunkhe.2023,Abbes.2024}, an open conjecture on the cuspidality is stated in \cite{Salunkhe.2023a}.

\begin{figure}
\begin{minipage}[c]{.2\linewidth}
        \centering
        \includegraphics[width=\linewidth]{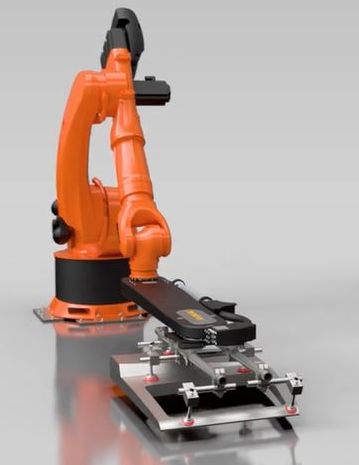}\\
        \textbf{a}
        %\label{fig:transpressor}
        %\captionof{figure}{KUKA Transpressor robot}
\end{minipage}\hfill
\begin{minipage}[c]{.55\linewidth}
        \centering
        \includegraphics[width=0.7\linewidth]{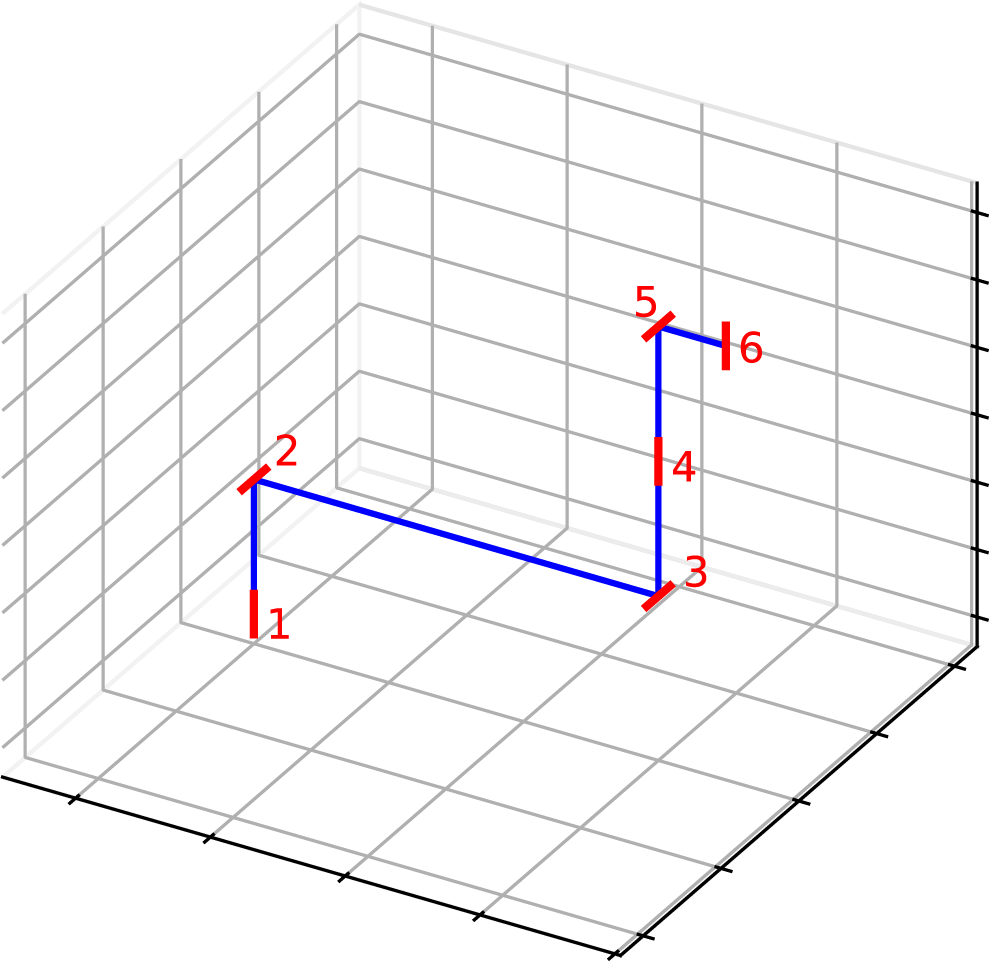}
        
        %\vspace{-5mm}
        \textbf{b}
        %\label{fig:homepos}
\end{minipage}\hfill
\begin{minipage}[c]{.2\linewidth}
    \centering
    %\captionof{table}{DH parameters for Transpressor-like robot as in \cite{WeissIMA2015}}
    \begin{tabular}{c|c|c|c}
        $i$  & $d_i$   &   $a_i$ & $\alpha_i$\\ 
    \hline
         1   & $d_1$   & 0  &  $\frac \pi 2$ \\
         2   & 0 & $a_2$ & 0 \\
         3   & 0   & 0 & $-\frac \pi 2$\\
         4   & $d_4$ & 0 & $\frac \pi 2$ \\
         5   & 0   & $a_5$  & $-\frac \pi 2$ \\
         6   & 0  & 0 & 0 
    \end{tabular}\\
    \vspace{3mm}
    \textbf{c}
    %\label{tab:DHparametersTranspressor}
\end{minipage}
\caption{ \textbf{a} The KUKA Transpressor, \textbf{b} our simplified version in home position (the red indication of axis $4$ is shifted along $z_4$ for visibility), and \textbf{c} the table of DH-parameters}
\label{figs:transpressor}
\end{figure}

We analyze a $6$R kinematic inspired by the KUKA Transpressor \cite{KUKATranspressor}. This robot is derived from a standard KUKA KR100P, but has axis 6 shifted away in parallel from the wrist center as shown in \Cref{figs:transpressor}. This makes the wrist non-spherical, and no analytical inverse kinematic solution is known. Our simplified version has intersections of axes $1$ and $2$, as well as axes $3$ and $4$, whereas the original robot has offsets between these axes. We investigate the IKS of this model and show its cuspidality. %As the structure of the robot is relatively simple, it appears to be a good candidate to analyze properties of cuspidal $6$R robots.

We proceed as follows: After introducing notation and describing the robot kinematics in \Cref{sec:preliminaries}, we extend the work of \cite{WeissIMA2015} in \Cref{sec:IKS} by providing a numerical approach for the general IKS and introducing another class of end effector poses with analytical solutions. This allows us to show the existence of a singularity-free change of IKS by analytical means in \Cref{sec:Singularity-free Change of Solution}. %To the best of the authors' knowledge all examples for cuspidal 6R robots rely on numerical evidence alone like in \cite{Carbonari.2023,Salunkhe.2023,Abbes.2024}. Of course these computations are convincing, but theoretical analysis might lead to deeper understanding.

\section{Robot Model}
\label{sec:preliminaries}

For any $T\in \hbox{SE}(3)$ we denote the rotational part by $R$ and the translation by $p$. The world coordinate system is set to align with the first axis $T_1 = T_{world}$. For DH-parameters $a_i,d_i$ and $\alpha_i$, we have
\begin{align*}
    T_{i+1} = T_i \cdot \Rotz{\theta_i} \cdot \Transl_z(d_i) \cdot \Transl_x(a_i) \cdot \Rotx{\alpha_i} \ .
%    T_{i+1} = T_i \cdot \Rotz{\theta_i} \cdot \Transl_z(a_i) \cdot \Transl_x(d_i) \cdot \Rotx{\alpha_i} \ .
\end{align*}
The end effector or target pose is denoted by $T_E$ and since we consider a $6$R manipulator, we have $T_E=T_7$ with the convention above. The coordinate axes of $T_i$ are denoted by $x_i,y_i$ and $z_i$, when referring to the world coordinate system we might omit the subscript.

\Cref{figs:transpressor}c shows the DH-parameters of the class of robots considered. For computations, we always set $d_1=0$, as different values can simply be compensated by moving the origin along the $z$-axis. Only in the visualizations we have $d_1 = 1$  and for the indication of joint axes we depicted axis $4$ shifted along $z_4$ to make the distinction between the axes easier (while actually the translational parts of $T_3$ and $T_4$ coincide). For visualizations and examples we usually set $a_2 = 3$, $d_4 = 2$ and $a_5 = 1/2$. The analytical and numerical methods for computing the inverse kinematic solutions described in \Cref{sec:IKS} can be applied for any choice of the non-zero parameters.

\section{Inverse Kinematic Solutions}
\label{sec:IKS}
After revisiting the kinematic of the regional $3$R chain composed of the inital three axes, we present a numerical approach to solve the inverse kinematic of the transpressor in general. 
%generally and describe analytical solutions for two families of special target poses.
%% new
We provide analytical solutions for poses with vertical $z_E$-axis, as well as for poses on the $z_1$-axis with a special orientation $R_E$. In both cases up to 16 solutions exist. For the latter class we will provide a singularity free change of solutions in \Cref{sec:Singularity-free Change of Solution}.
%% end new
\begin{figure}
\begin{minipage}[c]{.45\linewidth}
    \centering
    \includegraphics[width=\linewidth]{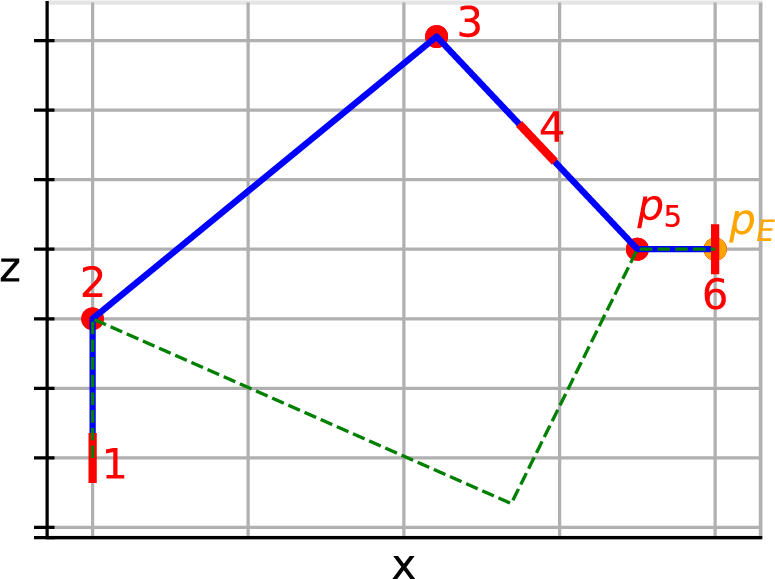}
    % \subcaption{Elbow up/down configurations with wrist outwards (everything lies flat in the $y=0$ plane)}
    % \label{subfig:elbow up/down}
    \textbf{a}
\end{minipage}
\begin{minipage}[c]{.45\linewidth}
    \centering
    \includegraphics[width=\linewidth]{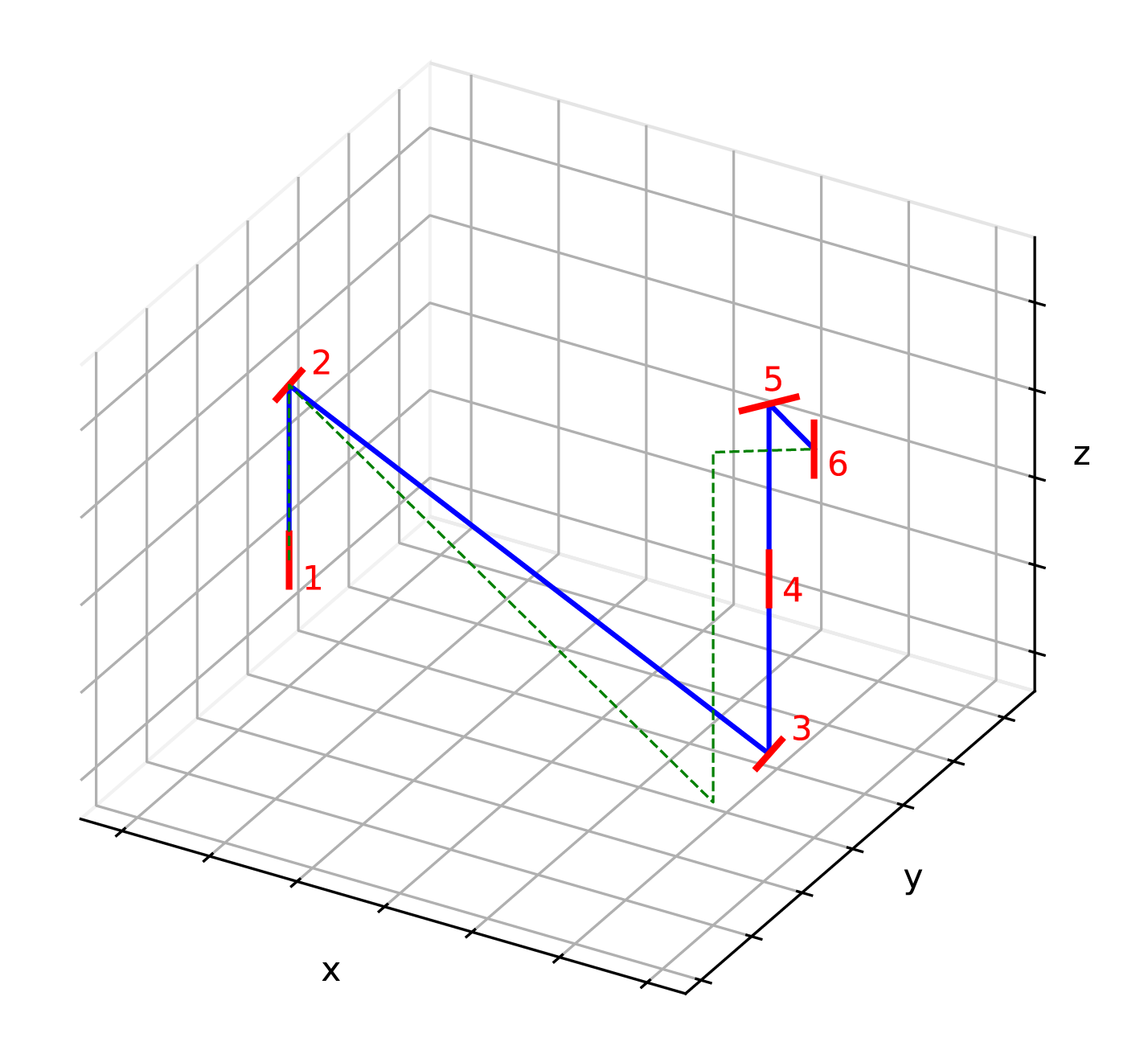}
    % \subcaption{Left/right configurations with elbow down}
    % \label{subfig:left/right} 
    \textbf{b}
\end{minipage}
\caption{Some IKS for a pose with vertical $z_E$-axis with \textbf{a} elbow up/down configurations, wrist outwards, and \textbf{b} left/right configurations, elbow down}
\label{fig:IKS_configurations}
\end{figure}

\subsection{Kinematic of the Initial $3$R-Chain}
\label{sec:3R_kinematic}
We consider the $3$R regional manipulator obtained by restricting to the first three joints of the robot with target position $p_5$, the fifth joint. The angle $\theta_4$ can be ignored or fixed, as it only affects the orientation of $T_5$. To compute the inverse kinematic with target $p_5=(p_{5,x}, p_{5,y}, p_{5,z})$ we set $\theta_1 = \arctan(p_{5,y} / p_{5,x})$ and then are able solve a planar $2$R chain for the other joints. This is illustrated in \Cref{fig:IKS_configurations}a and yields two solutions, commonly characterized as \emph{elbow up} or \emph{down}. Rotating $\theta_1$ by $\pi$ yields another two solutions. Explicitly, for any solution $(\theta_1,\theta_2,\theta_3)$ we can reach the same position with $(\theta_1 + \pi, -\theta_2 + \pi, -\theta_3 + \pi)$. This is called a \emph{shoulder flip}. In the reachable workspace, this $3$R-chain has generally exactly these four solutions, the only exception being singular configurations.% These occur when the two elbow configurations coincide (where one solution vanishes) and when the target position lies on the $z$-axis (where infinite solutions exist).

\begin{remark}
\label{rmk:elbow/shoulder flip}
    The shoulder flip can be generalized to the entire chain, one only has to correct the orientation of $T_5$ by rotating $\theta_4$ half a turn, i.e., the joint values $(\theta_1 + \pi, -\theta_2 + \pi, -\theta_3 + \pi, \theta_4 + \pi, \theta_5, \theta_6)$ yield the same pose as $(\theta_1,\theta_2,\theta_3,\theta_4,\theta_5,\theta_6)$.

    This does not work for the elbow change, as for $\theta_4 \neq 0,\pi$ we are generally not able achieve the same orientation at axis $5$ in both configurations.
\end{remark}

\subsection{General Approach}
\label{sec:general_iks}
\begin{figure}
    \centering
    \includegraphics[width=0.55\linewidth]{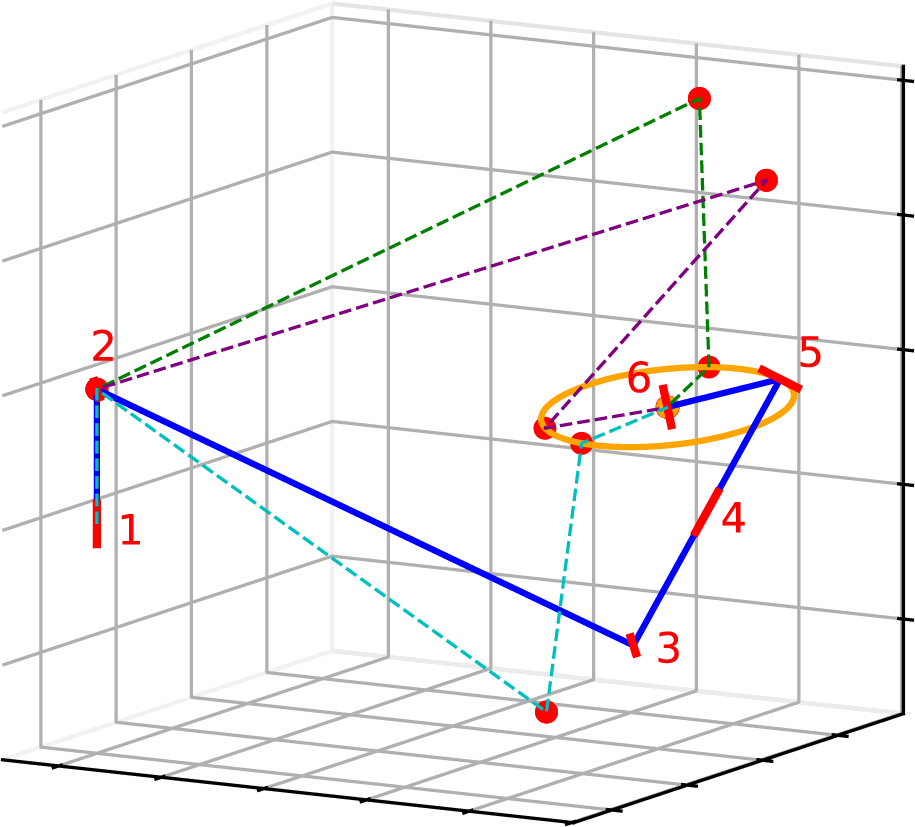}
    \caption{A pose with $4$ IKS, $2$ in each elbow configuration. Considering both shoulder configurations yields a total of $8$ IKS. For the solid blue configuration, we indicated the direction of the axes with red lines. Note that axis $5$ is tangential to the circle.}
    \label{fig:numerical_sols}
\end{figure}
Let $T_E$ be a target pose and $A$ the $x_E y_E$-plane. The last link has length $a_5$, therefore $p_5$ has to lie on a circle $S$ on $A$ with center $p_E$ and radius $a_5$ (depicted in orange  in \Cref{fig:numerical_sols}). Additionally, we need the axis $z_5$ to be tangential to $S$. Since axes $4$ and $5$ are orthogonal and intersect each other, we can equivalently ask for $z_4$ to be orthogonal to the tangent $t(p_5)$ to $S$ at $p_5$. As noted in \Cref{sec:3R_kinematic}, $p_5$ only depends on the first three axes. Therefore, the inverse kinematics problem can be reduced to regarding the initial $3$R-chain and investigating when it can reach $p_5\in S$ with $z_4 \perp t(p_5)$. By \Cref{rmk:elbow/shoulder flip} if one such solution is found, a second solution always exists via a shoulder flip, but the other elbow configuration has to be inspected separately.

This can be computed numerically, by discretizing the circle $S$, computing one IKS of the $3$R-chain on each point, and evaluating the inner product of $z_4$ and $t(p_5)$. While increasing the number of sample points results in higher accuracy, the number of solutions is usually already correct with $n=50$, as this is only critical when distinct IKS only differ marginally (e.g. near a singularity). %Gradient-based iterative numerical algorithms might also be used but depend on initial values, so we might miss one solution.

\subsection{Poses with Vertical $z_E$-Axis}
\label{sec:vert_z}

In this section we provide an analytical method to compute all IKS for target poses with $z_E \parallel z_1$. For simplicity, we may even assume $T_E$ to be a pure translation, as we can always achieve this by an appropriate choice of $\theta_6$. Similarly, we can restrict to poses on the $xz$-half-plane with positive $x$-values, and get the other positions by rotating about $\theta_1$.

The fundamental observation for these considerations is the following: The plane $A$ described in \Cref{sec:general_iks} is now parallel to the $x_1y_1$-plane. For any configuration $(\theta_1, \theta_2, \theta_3)$ with $p_5 \in A$, the rotation of joint $4$ can be regarded as a circle around $p_5$, which the $z_5$-unit vector traces. This circle intersects $A$ exactly twice, at the values $\theta_4 = 0$ and $\theta_4 = \pi$. This means these poses can generally only be reached if all links align (projected along the $z$-axis). For fixed elbow and shoulder configuration of the $3$R-chain, this yields up to two solutions with the wrist either facing \emph{out-} or \emph{inwards}. In total, this results in up to $8$ IKS, for which the remaining joint values are computed by solving the $3$R inverse kinematics for $p_5 = (p_{E,x} \pm a_5, 0, p_{E,z})$. The wrist out configuration is shown in \Cref{fig:IKS_configurations}a.

There exists an additional possibility to reach these poses if there is a point on $S$ that can be achieved with $z_4 \parallel z_1$. Then the $z_5$-axis lies in $A$ for any $\theta_4$ and we can choose a suitable value to get $z_4 \perp t(p_5)$. If we can reach a target pose that way with $\theta_4$ as the fourth joint value, we can also reach it from a similar pose with $\theta_4' = - \theta_4$. Therefore, these solutions always appear in pairs of a \textit{left} and \textit{right} configuration as depicted in \Cref{fig:IKS_configurations}b. In this case we have to be more careful about the configuration of the $3$R-chain, as usually changing the elbow configuration changes the direction of $z_4$. Thus we need to separately check both possibilities and it might be that in neither, both or only in one elbow configuration the special solutions can be obtained. 

Explicitly, consider a target pose as above, with orientation identical to the orientation of $T_1$, and translational part $p_E=(p_{E,x},0,p_{E,z})$ with $p_{E,x} \geq 0$. The condition $z_4 \parallel z_1$ is satisfied if and only if $\theta_3 = \pi-\theta_2$ in the elbow up configuration and $\theta_3 = -\theta_2$ with elbow down. If $p_{5,z}=p_{E,z}$ can be achieved, then there exists exactly one pair $(\theta_2, \theta_3)$ for each shoulder configuration. Fixing these, we consider the distance $\rho_5$ with $\rho_5^2 = p_{5,x}^2 + p_{5,y}^2$. Now we can find a solution if and only if $|\rho_5 - \rho_E| \leq a_5$. In the case of equality, both solutions coincide with one of the solutions with wrist out- or inwards. Given a strict inequality, we obtain exactly two solutions, and readily compute the remaining angles to approach $p_E$ from two sides. Considering both elbow and shoulder configurations, we obtain up to $8$ additional solutions. Analytic formulas are possible.

\subsection{Poses on $z_1$-Axis}
\label{sec:z-axis}

\begin{figure}
    \centering
    \includegraphics[width=0.45\linewidth]{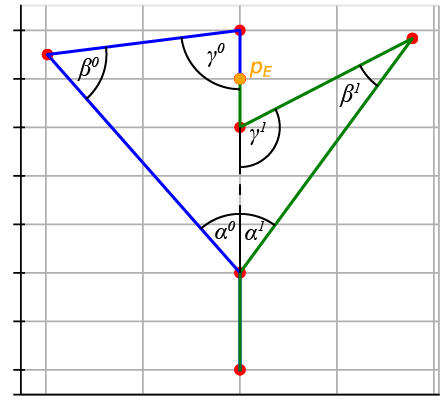}
    \caption{Two configurations to reach a pose on the $z$-axis: left configuration with elbow up and no wrist flip (blue), and right configuration with elbow up and wrist flipped (green)}
    \label{fig:triangle-construction}
\end{figure}

For the subsequent analysis, we recall the second family of poses with analytical inverse kinematics solutions. For a more detailed description we refer to \cite{WeissIMA2015}. The results can be applied to target poses of the form
\begin{align}
\label{eq:endeffectorpose}
    T_E = 
    \begingroup % keep the change local
    \setlength\arraycolsep{6pt}
    \begin{pmatrix}
        0 & 0 & 1 & 0\\
        0 & 1 & 0 & 0\\
        -1 & 0 & 0 & p_{E,z} \\
        0 & 0 & 0 & 1
    \end{pmatrix} \ .
    \endgroup
\end{align}
With these, we can obtain solutions either by aligning the wrist extension with the $z_1$-axis (\textit{left/right configurations}) or with the $z_4$-axis (\textit{back/front configurations}). Switching between left and right (or back and front) is achieved by a shoulder flip. Each of these again has two possible directions of the wrist (\textit{flipped or not}). Finally, all of these can be reached in both elbow configurations, yielding up to $16$ solutions. Two of the four configurations are shown in \Cref{fig:triangle-construction}. Formulas are given in \cite{WeissIMA2015}.

\subsection{Number of IKS}

For both analytical approaches, the number of IKS can be $0, 4, 8, 12$ or $16$, depending on the target pose. Applying the numerical algorithm indicates that the same holds generally, but grouping these solutions remains an open challenge.

\section{Singularity-free Change of Solution}
\label{sec:Singularity-free Change of Solution}

We now provide a singularity-free motion between two solutions $\theta^{0}, \theta^{1}\in\R^6$, reaching the same end effector pose.
% Implications for the number of aspects:
% vielleicht kommen wir ohne die Erklärung samt figure aus
% z = 4 wählen statt z=2, sonst z+a5 = d4+a5, zufällige Symmetrie
We regard the configurations depicted in \Cref{fig:triangle-construction}, the joint values are provided in \Cref{tab:16poses}.

% \begin{eqnarray*}
% 	\Delta(d_4, z+a_5, a_2) & \mapsto& (\alpha^{1}, \beta^{1}, \gamma^{1}) 
% 	\qquad \hbox{ left and right, extension down, no wrist flip }
% 	\\
% 	% \Delta(d_4 + a_5, z, a_2) & \mapsto& (\alpha^{2}, \beta^{2}, \gamma^{2})
% 	% \qquad \hbox{ back and front, no wrist flip }
% 	% \\
% 	\Delta(d_4, z-a_5, a_2) & \mapsto& (\alpha^{3}, \beta^{3}, \gamma^{3})
% 	\qquad \hbox{ left and right, extension up, wrist flip }
% 	% \\
% 	% \Delta(d_4-a_5, z, a_2) & \mapsto& (\alpha^{4}, \beta^{4}, \gamma^{4})
% 	% \qquad \hbox{ back and front, wrist flip }
% \end{eqnarray*}

% \begin{figure}
% \centerline{
% \includegraphics[width=60mm]{images/xplanetrans1a.png} 
% \includegraphics[width=60mm]{images/xplanetrans9a.png} 
% }
%  \caption{Left and right configurations of the robot}
%   \label{fig:triangleleftright}
% \end{figure}
% \begin{figure}
% \centerline{
% \includegraphics[width=60mm]{images/xplanetrans5a.png} 
% \includegraphics[width=60mm]{images/xplanetrans13a.png} 
% }
%  \caption{Front and back configurations of the robot}
%   \label{fig:trianglefrontback}
% \end{figure}

As an example we prove that the linear path $\theta(\lambda) = \theta^{0} + \lambda (\theta^{1}-\theta^{0})$, $\lambda\in[0,1]$ is without singularity for a wide range of parameters.

\begin{table}
  \caption{The two configurations investigated}
\centerline{
\begin{tabular}{c||c|c|c|c|c|c||l}
  & $\axis_1$ & $\axis_2$ & $\axis_3$ & $\axis_4$ & $\axis_5$ & $\axis_6$ & description \\
\hline
$\theta^0$	&   $0$   &  $\frac \pi 2 -\alpha^0$    &  $\frac \pi 2 -\beta^0$   &   $0$     &   $-\frac \pi 2 -\gamma^0$    &  $0$ &  shoulder left, elbow up, no wrist flip \\
$\theta^1$ &  $\pi $ & $ \frac  \pi 2 -\alpha^1 $ &  $ \frac  \pi 2 -\beta^1 $ &   $0$  &  $ \frac  \pi 2 -\gamma^1 $ & $ \pi$ & shoulder right, elbow up, wrist flipped \\

\end{tabular}
}
  \label{tab:16poses}
\end{table}

The determinant of the Jacobian is
\begin{equation}
\label{eq:determinant}
% \det (J(\theta)) = 
- a_2 c_3 ((( s_3 d_4^2- a_2 d_4) s_5 + a_5 (a_2 c_4^2 + d_4 s_3 - a_2)) c_2
+ s_2 d_4 c_3 (s_5 d_4 + a_5))  \ .
\end{equation}
We abbreviate $s_i = \sin(\theta_i)$ and $c_i = \cos(\theta_i)$.
% The idea for the proof of singularity-free paths is: The 16 solutions exist for any $a_5\neq 0$, but for small $a_5$ the triangles and their angles are approximately the same in all four cases. This allows estimates for the signs of the terms in (\ref{eq:determinant}).
The value $\theta_4(\lambda) = 0$ is constant and therefore $c_4=1$, simplifying (\ref{eq:determinant}) to
\begin{eqnarray}
 & & 
 - a_2 c_3 ((( s_3 d_4^2- a_2 d_4) s_5 + a_5 d_4 s_3)) c_2
+ s_2 d_4 c_3 (s_5 d_4 + a_5))  
\nonumber
\\
 & = &
 - a_2 d_4 c_3 ((s_3 d_4 - a_2) s_5 + a_5 \sin(\theta_2+\theta_3) +  s_2 c_3 s_5 d_4)
 \nonumber
\\
 & = &
 - a_2 d_4 c_3 (((s_3 +s_2 c_3) d_4 - a_2) s_5 + a_5 \sin(\theta_2+\theta_3))
\label{eq:det-factored}
\end{eqnarray}
The factor $c_3$ in front of the bracket does not change its sign if $0 < \beta^k < \pi/2$ for $k=0,1$. A sufficient condition is that the triangle $\Delta(d_4, z+a_5, a_2)$ for configuration $\theta^0$ has an acute angle $\beta^0$, then also $\Delta(d_4, z-a_5, a_2)$ for configuration $\theta^1$ has an acute angle $\beta^1$ for monotonicity reasons. To achieve this, $|p_{E,z}\pm a_5|$ must be small compared to $a_2$ and $d_4$.

The bracket contains a summand with factor $\sin(\theta_2+\theta_3)$ and one with factor $\sin(\theta_5)$. We give sufficient conditions for both summands to be positive. 

We have $\alpha^k+\beta^k+\gamma^k=\pi$ for both $k=0,1$. From \Cref{tab:16poses} we see that $\theta_2+\theta_3 = \pi - \alpha^k - \beta^k = \gamma^k$ for both poses $0$ and $1$. Moreover, $\theta_5^0 = -\pi/2 - \gamma^0$ and $\theta_5^{1} = \pi/2 - \gamma^1$. 
%In both cases the difference between $\theta_5$ and $\theta_2+\theta_3$ is $\pi/2$, so one of the $\sin$ values will be close to 0 if the other is close to 1 in absolute value.
In both cases $\sin(\theta_5) = \pm \cos(\theta_2+\theta_3)$, so one of the $\sin$ values will be close to $0$ if the other is close to $1$ in absolute value.

Choosing $\gamma$ close to $\pi/2$, we see that $\sin(\theta_2(\lambda)+\theta_3(\lambda))$ has positive values near $1$ in the interval between $\gamma^0$ and $\gamma^1$ because $\theta_2(\lambda)+\theta_3(\lambda) = \gamma^0+\lambda(\gamma^1 - \gamma^0)$ holds on the whole connection. 

This suggests choosing $p_{z,E}$ such that the triangle $\Delta(d_4, p_{E,z}, a_2)$ with sides $p_{z,E}$, $a_2$ and $d_4$ has a right angle at $p_E$. With this choice, we observe in \Cref{fig:triangle-construction} that $0 < \gamma^0 < \pi/2$ and $\pi/2 < \gamma^1 < \pi$. This translates into $-\pi < \theta_5^0< 0$ and $-\pi/2 < \theta_5^1<0$. So we have $\sin(\theta_5(\lambda))< 0$ for all $\lambda\in[0,1]$. The term before $\sin(\theta_5)$ can be made negative if we require $a_2> 2 d_4$. We conclude that the bracket term in (\ref{eq:det-factored}) is positive.

Numerical computations show that the estimates are quite conservative. Our usual example values $a_2=3, d_4=2, a_5=1/2$ violate the condition $sin(\theta_5(\lambda))< 0$ slightly at the boundaries of the interval but the determinant still does not change sign. 
% For many other connections among the $16$ IKS similar arguments are possible, but we cannot give a common proof for all endpoints $\theta^i$ and $\theta^j$. 
% to do: Verbindung Koordinatenweise aufbauen. also zuerst \theta_2 ändern, alles andere konstant lassen

% to do: wenn wir noch eine Seite hätten: es gibt "einfach abschätzbare" Verbindungen wie 1 -> 11, und solche, die hart an 0 herangehen, und darum 2 Linien im Achsraum für eine singularitätsf reie Verbindung brauchen. Meine Bilder:
% \includegraphics[width=0.45\textwidth]{images/allCandidates16_z2.jpg}
% \includegraphics[width=0.45\textwidth]{images/candidates1_z2.png}

The above provides an analytical way of showing cuspidality for a range of different DH-parameters. Generalizing this to other solution pairs and different DH-parameters can get cumbersome. For the parameter values $a_2=3, d_4=2$ and $a_5=1/2$ we investigated these solution changes also numerically. Considering again the solutions of \Cref{sec:z-axis}, one can group the $16$ IKS into $4$ classes of $4$ solutions such that within each class, the sign of each of the two factors in the Jacobi Determinant does not change. While there exist more elaborate approaches to detect connectedness \cite{capcoRobotsComputerAlgebra2020}, we discovered that just a small adjustment to a linear motion (e.g. by translating the path slightly along $\theta_1$) leads to a singularity-free path between any two solutions in the same class. This means all IKS of one class are in the same aspect, i.e., a singularity-free region in joint space.

\section{Conclusion and Outlook}

We provided algorithms to compute the IKS of a class of robots with up to $16$ solutions. Moreover, the cuspidality property for a range of parameters was established and proved analytically.

The results suggest that further investigation might prove fruitful, but even this simple architecture shows that the analysis of $6$R cuspidal robots is challenging: An analytical solution for the entire workspace would be desirable, but the results of \Cref{sec:vert_z,sec:z-axis} do not easily generalize. 

It would be useful to classify the configurations into singularity-free groups by their parameter values and establishing partitions of work- and joint space where movement avoiding singular configurations is possible. There, studying the singular variety and symmetries in the IKS could be helpful. Understanding these kinematic properties is essential to safe and efficient motion planning.

%The symmetry of the analytical solutions in \Cref{sec:vert_z,sec:z-axis} does not easily generalize. The solutions appear in pairs, but classifying them into different configurations does require further insight. Here, studying the singularities can be useful. Understanding these kinematic properties is essential to safe and efficient motion planning.

%including generalizations: offsets in kinematics work, TCP near $x$ and $z$-axis works


\end{document}